\documentclass[runningheads]{llncs}
\usepackage{amsmath}
\usepackage[misc,geometry]{ifsym}
\usepackage[T1]{fontenc}
\usepackage{tabularx} % Add this line to your preamble
\usepackage{graphicx}
\usepackage{float}
\usepackage{hyperref}

\usepackage[numbers]{natbib}
\usepackage{array}
\usepackage{tabularx}
\usepackage{multirow} % Add this package for multirow functionality
\usepackage{float} % For the H option in tables

\begin{document}

\title{TI-JEPA: An Innovative Energy-based Joint Embedding Strategy for Text-Image Multimodal Systems}

\titlerunning{TI-JEPA}
% If the paper title is too long for the running head, you can set
% an abbreviated paper title here
%

\author{Khang H. N. Vo\inst{1,2\star} \and
Duc P. T. Nguyen \inst{1,2\star} \and
Thong T. Nguyen \inst{3} \and
(\Letter) Tho T. Quan \inst{1,2}}
% TODO: Tho T. Quan
\renewcommand{\thefootnote}{\fnsymbol{footnote}}
\footnotetext[1]{The two authors contributed  equally to this paper.}
\authorrunning{Khang H. N. Vo, Duc P. T. Nguyen, et al.}
% First names are abbreviated in the running head.
% If there are more than two authors, 'et al.' is used.
%
\institute{URA Research Group, Faculty of Computer Science and Engineering, Ho Chi Minh City University of
 Technology (HCMUT), Ho Chi Minh City, Viet Nam \and Vietnam National University Ho Chi Minh City, Ho Chi Minh City, Viet Nam \and National University of Singapore, Singapore \\ \email{qttho@hcmut.edu.vn}}
% TODO: Vietnam National University Ho Chi Minh City
\maketitle              % typeset the header of the contribution
\begin{abstract}
% The abstract should briefly summarize the contents of the paper in
% 150--250 words.

This paper focuses on multimodal alignment within the realm of Artificial Intelligence, particularly in text and image modalities. The semantic gap between the textual and visual modality poses a discrepancy problem towards the effectiveness of multi-modalities fusion. Therefore, we introduce Text-Image Joint Embedding Predictive Architecture (TI-JEPA), an innovative pre-training strategy that leverages energy-based model (EBM) framework to capture complex cross-modal relationships. TI-JEPA combines the flexibility of EBM in self-supervised learning to facilitate the compatibility between textual and visual elements. Through extensive experiments across multiple benchmarks, we demonstrate that TI-JEPA achieves state-of-the-art performance on multimodal sentiment analysis task (and potentially on a wide range of multimodal-based tasks, such as Visual Question Answering), outperforming existing pre-training methodologies. Our findings highlight the potential of using energy-based framework in advancing multimodal fusion and suggest significant improvements for downstream applications.
% TODO: Bỏ mấy cái text-image
% 1 2 câu: Các cách tiếp cận cũ đang có vấn đề/đang dùng những thứ gì -> Chúng tôi dùng energy & contrastive
% Ko lặp lại 1 từ/tên quá nhiều
% So sánh thí nghiệm cho thấy outperform...

% TODO: Them Joint Embedding
\end{abstract}

\section{Introduction}

In the era of Artificial Intelligence, the ability to process and understand information from multiple modalities simultaneously has become increasingly crucial \cite{Zhao_Li_Xu_2022,zhao2022cognitive}. \textit{Multimodal fusion}, the process of integrating information from various sensory inputs to form a coherent understanding, stands at the forefront of this challenge. Among the myriad of multimodal tasks, text-image alignment has emerged as a fundamental problem with far-reaching applications in areas such as visual question answering, image captioning, and cross-modal retrieval \cite{VBFusion2022}.
% TODO: Reference cho câu thứ nhất
% Note: tốt nhất 10-15 references cho 1 paper
% Lần đầu nhắc tới keyword phải in nghiêng

Despite significant advancements in natural language processing and computer vision independently, aligning these two modalities remains a formidable challenge \cite{Nguyen2024KDMCSEKD,nguyen-etal-2022-adaptive}. The semantic gap between the continuous, high-dimensional space of images and the discrete, symbolic nature of text poses substantial difficulties. Traditional approaches often struggle to capture the latent relationships between visual and textual elements, leading to sub-optimal performance in downstream tasks.
% Reference phải bắt đầu từ 1

Different from traditional approaches, Energy-based models (EBMs) \cite{LeCun2006ATO,nguyen-etal-2023-demaformer} have recently shown promise in various machine learning domains due to their flexibility and ability to capture complex dependencies. By learning a scalar energy function that associates low energy with correct configurations and high energy with incorrect ones, EBMs offer a natural framework for modeling the compatibility between different modalities. However, their application to text-image alignment has been limited, leaving room for innovation in this critical area.

Therefore, we propose Text-Image Joint Embedding Predictive Architecture (TI-JEPA), a novel pre-training approach for text-image alignment which leverages EBM concepts to capture complex dependencies. Extensive experiments demonstrate that TI-JEPA achieves state-of-the-art performance on a wide range of text-image alignment benchmarks, demonstrating its effectiveness and versatility.

To sum up, the main contributions of this work are as follows.

\begin{itemize}
    \item We propose TI-JEPA as a new pre-training strategy for text-image alignment, leveraging EBM concepts to capture complex cross-modal relationships. TI-JEPA learns strong relationships between text and image representations.
    
    \item We introduce TI-JEPA as a flexible and dynamic framework for multimodal training, capable of producing a robust multimodal encoder. This framework can effectively capture the relationships between text and image while utilizing pre-trained encoders for enhanced performance and scalability.
    
    \item We demonstrate that TI-JEPA is competitive with state-of-the-art (SOTA) pre-training methodologies, showing improved performance in both accuracy and F1-score on multimodal sentiment analysis task.
\end{itemize}

\section{Related Works}

\subsection{Multimodal Fusion}

Multimodal fusion has been a growing area of interest in machine learning \cite{nguyen2023visionandlanguagepretraining,nguyen2024encodingcontrollingglobalsemantics}. Early works \cite{NEURIPS2019c74d97b0,pmlr-v37-xuc15,Liu2018EfficientLM} focused on feature-level fusion, combining representations from different modalities using simple concatenation or averaging.

More advanced techniques have since emerged. Xu et al. \cite{Liu2018EfficientLM} introduced attention-based fusion, allowing models to dynamically focus on relevant features across modalities. They also proposed tensor-based methods for capturing higher-order interactions between modalities. Lu et al. \cite{NEURIPS2019c74d97b0} explored the use of Transformer architecture for multimodal fusion, leveraging their ability to model long-range dependencies across different data types.

\subsection{Text-Image Alignment}

Text-image alignment has seen significant advancements in recent years \cite{NIPS20137cce53cf,nguyen2023improvingmultimodalsentimentanalysis,radford2021learningtransferablevisualmodels}. Frome et al. \cite{NIPS20137cce53cf} introduced the concept of visual-semantic embeddings in DeViSE, learning a joint space where semantically similar text and images are close to each other.
While Radford et al. \cite{radford2021learningtransferablevisualmodels} demonstrated the power of contrastive learning with large-scale image-text data for text-image alignment with the CLIP model. This approach has since been extended by methods like ALIGN \cite{Jia2021ScalingUV}, which further improved performance through larger-scale training.
More recently, Li et al. \cite{li-etal-2021-unimo} proposed UNIMO, a unified framework for text-image pretraining, incorporating multiple pretext tasks to learn robust representations.

\subsection{Self-Supervised Learning in Multimodal Domain}

Self-supervised learning has emerged as a powerful paradigm for learning representations without relying on explicit labels, significantly advancing various fields of artificial intelligence \cite{NEURIPS2018c4616f5a,li-etal-2021-unimo,NEURIPS2019c74d97b0,10.1145/3581783.3612248,nguyen-etal-2024-video,nguyen2024topicmodelingmultiobjectivecontrastive, nguyen2024motion, nguyen2024meta, nguyen2024multi, nguyen2021contrastive, wu2024modeling, nguyen2025enhancing}. This approach has shown particular promise in the multimodal domain, where it can leverage the rich, complementary information present across different modalities. The application of self-supervised learning to text-image tasks has led to significant advancements in multimodal understanding:

\begin{enumerate}
    \item \textbf{Masked Language Modeling for Multimodal Inputs}: Lu et al. \cite{NEURIPS2019c74d97b0} adapted the successful masked language modeling technique from natural language processing to handle multimodal inputs. Their approach allows the model to learn joint representations of text and images by predicting masked tokens in the presence of both modalities.

    \item \textbf{Unified Masked Modeling}: Building upon this concept, Li et al. \cite{li-etal-2021-unimo} proposed UNIMO, a unified framework that applies masked modeling objectives to both text and image modalities simultaneously. This approach enables more robust and versatile multimodal representations, as the model learns to understand and generate content across modalities in a unified manner.
\end{enumerate}
These advancements in self-supervised learning for multimodal tasks have paved the way for improved performance in various applications, including visual question answering, image captioning, and cross-modal retrieval \cite{nguyen-etal-2024-video,nguyen2024topicmodelingmultiobjectivecontrastive}. The ability to learn from unlabeled multimodal data has also opened up new possibilities for processing and understanding the vast amounts of unstructured multimodal content available on the internet \cite{10.1145/3581783.3612248}.
\subsection{Energy-Based Models (EBMs) and Joint-Embedding Predictive Architecture (JEPA)}

EBMs are a class of probabilistic models that define a probability distribution over the input space by associating an "energy" value with each possible input \cite{Ou2024EnergyBasedMW}. The energy function is typically parameterized by a neural network, and the probability of an input is inversely proportional to its energy. This formulation allows EBMs to model complex, non-linear relationships in the data, making them well-suited for tasks such as speech and language processing.

EBMs have recently regained attention in the machine learning community. LeCun et al. \cite{LeCun2006ATO} provided a comprehensive overview of EBMs and their applications. Assran et al. \cite{assran2023selfsupervisedlearningimagesjointembedding} introduced the Image-based Joint-Embedding Predictive Architecture (I-JEPA), a non-generative self-supervised learning approach that focuses on predicting representations of various target blocks within the same image. This method effectively scales with Vision Transformers and achieves impressive downstream performance across tasks like linear classification and object counting, further showcasing the versatility of self-supervised learning architectures.

\section{Our Approach}

\begin{figure}
    \centering
    \includegraphics[width=\linewidth]{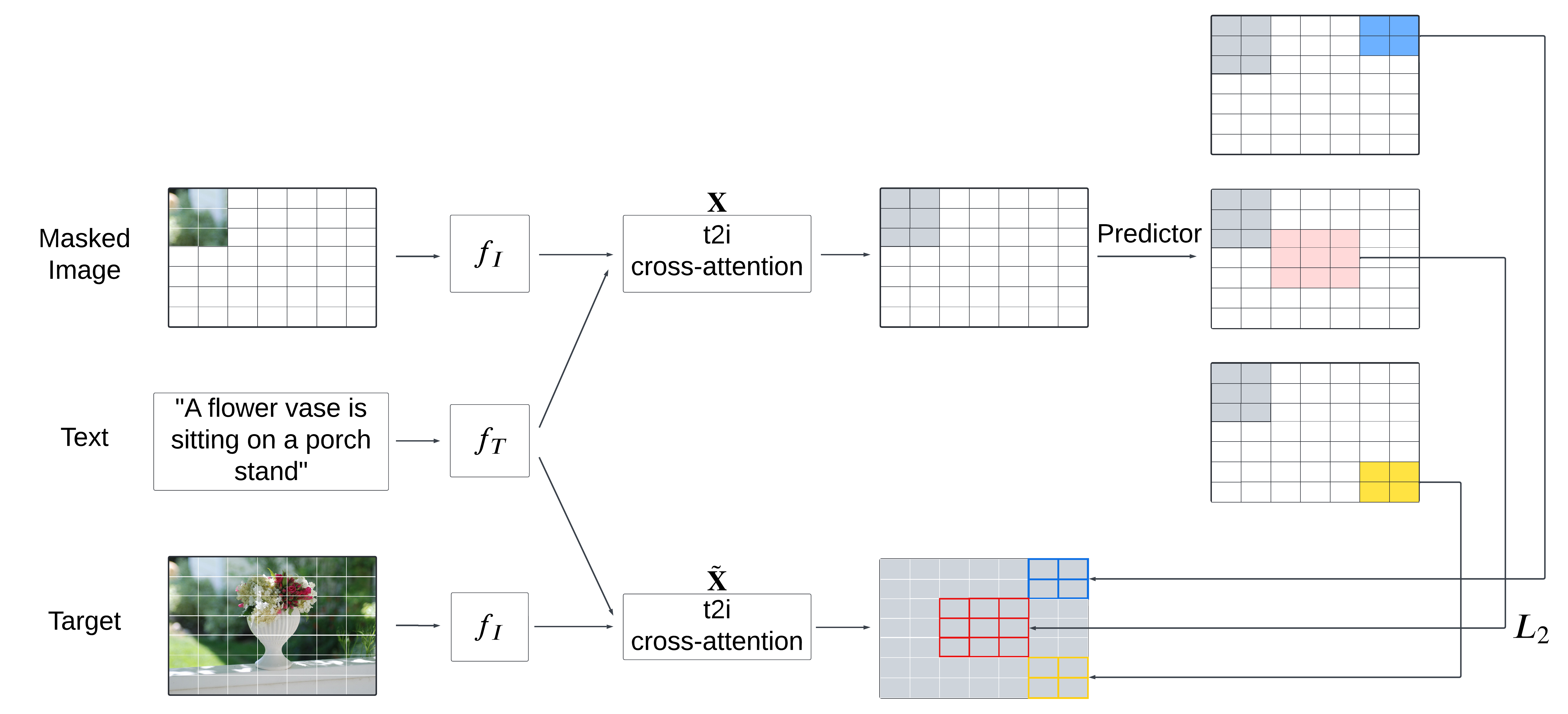}
    \caption{The proposed TI-JEPA architecture, where cross-attention between text and image encodings is leveraged to predict masked patches.}
    \label{fig:arch}
\end{figure}

The proposed TI-JEPA architecture integrates cross-attention mechanisms to effectively align textual and visual information for predicting masked image patches, which is demonstrated in Figure \ref{fig:arch}. Before getting into details, denote $I$ as the original image, $I_{\text{context}}$ as the same image but after performing context masking, and its caption as $T$. The high-level aspect of our architecture can be described in smaller components as below:

\begin{itemize}
    \item Image encoder $f_I$ processes on \textit{full} image $I$ and masked image $I_{\text{context}}$, generate its full embedding representation and masked context parts (context block) representation.
    
    \item Text encoder $f_T$ converts image description $T$ into a dense representation that captures semantic information.
    
    \item We employs two blocks of text-to-image (\texttt{t2i}) cross attention, namely block $\mathbf{X}$ and block $\tilde{\mathbf{X}}$, to align the encoded text features with the visual features from the image encoder.
    
    \item The output of the \texttt{t2i} cross-attention block $\mathbf{X}$ is passed through a predictor $g_{\phi}$, which generates the final predictions for the representations of the target patches.
\end{itemize}

\subsection{Creating Target and Context Blocks}

To create the target and context representations, we designed an approach which consists of two stages: Target representation creation step and context representation creation step.

\noindent\textbf{Target Representation Creation.} The first step involves creating target representations for the blocks to be predicted. Particularly, the target representations correspond to a combination of embedding vectors from the text and the target image patches. Given an input image $I$, we divide it into $N$ non-overlapping patches, which are passed through the image encoder $f_{I}$ to obtain corresponding representations $\mathbf{s}_I = \{\mathbf{s}_{I_1}, \mathbf{s}_{I_2}, \dots, \mathbf{s}_{I_N}\}$, where $\mathbf{s}_{I_k}$ is the representation for the $k^{th}$ patch. The paired text $T$ will be divided into $L$ tokens, which are passed through the target encoder $f_{T}$ to obtain corresponding representations $\mathbf{s}_T = \{\mathbf{s}_{T_1}, \mathbf{s}_{T_2}, \dots, \mathbf{s}_{T_L}\}$, where $\mathbf{s}_{T_k}$ is the representation for the $k^{th}$ token of the text. Finally they will go through the cross-attention to generate the final target representations $\mathbf{s}_y = \tilde{\mathbf{X}}(\mathbf{s}_T, \mathbf{s}_I)$. To obtain the targets for the loss, from those representations we sample $M$ blocks, which are going to include the patches that needs to be predicted. We denote the mask corresponding to the $i^{th}$ block with $B_i$, and its patch-level representation with $\mathbf{s}_y(i)=\{\mathbf{s}_{y_j}\}_{j \in B_i}$.

\noindent\textbf{Context Representation Creation.} For context representations, we randomly sample a single block $I_x$ from the image $I$ with a random scale. We denote the mask associated with this context block as $B_x$. To prevent trivial predictions, we remove any regions that overlap between the context blocks and the target blocks created earlier. The masked context block are also passed through the image encoder $f_{I}$, and through the context cross-attention module $\mathbf{X}$ with the encoded text to obtain corresponding representations $\mathbf{s}_x=\{\mathbf{s}_{x_i}\}_{i \in B_x}$.

\subsection{Output Prediction}

The output of the attention module forms the context vectors, which serve as the input to the predictor. More formally, given the output $\mathbf{s}_x$ from the attention module, the goal is to predict the representations for the $M$ target embeddings: $\mathbf{s}_y(1), \mathbf{s}_y(2), \dots, \mathbf{s}_y(M)$. For each target embedding $\mathbf{s}_y(i)$, corresponding to the target mask $B_i$, the predictor takes as input the attention output $s_x$ and a set of mask tokens $\{m_j\}_{j \in B_i}$ for each patch that needs to be predicted. The predictor then outputs a prediction $\{\hat{\mathbf{s}}_{y_j}\}_{j \in B_j} = g_{\phi}(\mathbf{s}_x, \{m_j\}_{j \in B_i})$. The mask tokens are parameterized by a shared learnable vector with added positional encoding.

% TODO: ???? cai nay yapping cai gi v
We obtain the final predictions $\hat{\mathbf{s}}_y(1), \hat{\mathbf{s}}_y(2), \dots, \hat{\mathbf{s}}_y(M)$ by applying the predictor $M$ times, each time we rely on the mask tokens for the corresponding target-block locations.

\subsection{Objective Function}

Our objective function is to minimize the average $L_1$ distance between the predicted representations $\hat{\mathbf{s}}_y(i)$ and the true target representations $\mathbf{s}_y(i)$:
    
    \begin{equation*}
        L_{\text{P}} = \frac{1}{M} \sum_{i=1}^{M} D(\hat{\mathbf{s}}_y(i), \mathbf{s}_y(i)) = \frac{1}{M} \sum_{i=1}^{M} \sum_{j \in B_i} \| \hat{\mathbf{s}}_{y_j} - \mathbf{s}_{y_j} \|^2_2
    \end{equation*}

\section{Training TI-JEPA}

\subsection{Dataset}

Our experimental setup was designed to thoroughly evaluate the performance and capabilities of the TI-JEPA model. For training, we utilized the Microsoft COCO 2017 dataset \cite{Lin2014MicrosoftCC}, specifically the training set, which comprises over 118,000 image-text pairs. The COCO dataset was chosen for its exceptional diversity in both visual content and descriptive captions, providing a comprehensive foundation for multimodal training. Since each image in the dataset is associated with multiple captions, we randomly selected one caption per image to create a unique text-image pair for each record. This process ensures that our dataset contains distinguishable and meaningful text-image pairs, enabling more effective multimodal learning.

\subsection{Experiment and Hyperparameter Configurations}

All experiments were conducted on two NVIDIA GeForce GTX 1080 Ti GPUs. The training process spanned 300 epochs, with the largest model requiring approximately 188 hours on our hardware setup.

\textbf{Encoder Modules:} For the image encoder, we utilized a checkpoint pretrained with the ViT-H architecture, using a $16 \times 16$ patch size and $224 \times 224$ resolution, trained for 300 epochs as part of the I-JEPA model\footnote{Checkpoint available here: \url{https://github.com/facebookresearch/ijepa?tab=readme-ov-file\#pretrained-models}}. The text encoder is based on \texttt{gte-base-en-v1.5}, a model developed by the Institute for Intelligent Computing at Alibaba Group, using the \texttt{transformer++} backbone (BERT + RoPE + GLU). It supports a context length of up to 8192 tokens with a text embedding dimension of 768.

\textbf{Cross-Attention Module Variants:} The \texttt{t2i} cross-attention module consists of multiple blocks, each containing a self-attention layer, a cross-attention layer with text representations, and a multilayer perceptron (MLP) layer, all with residual connections. We configured three variants of the module: Small, Medium, and Large, as detailed in Table \ref{tab:cross} below.

\begin{table}[ht] 
\centering 
\caption{Cross-attention module variants} \label{tab:cross} 
\begin{tabular}{|l|c|c|c|c|} 
\hline \textbf{Model} & \textbf{Layers} & \textbf{Heads} & \textbf{Hidden Size} & \textbf{Params} \\ \hline 
Small & $4$ & $8$ & $768$ & $39$M\\ \hline 
Medium & $6$ & $10$ & $768$ & $58$M\\ \hline 
Large & $8$ & $12$ & $1024$ & $131$M\\ \hline 
\end{tabular} \end{table}

\textbf{Predictor Module:} The predictor is inherited from the original I-JEPA predictor, which is a shallow vision transformer with a depth of 12 layers and 12 attention heads per layer.

\textbf{Training configs:} We utilized a learning rate of $0.001$ and a momentum scheduler with an exponential moving average (EMA) ranging between $0.996$ and $1.0$. The model was trained with a batch size of $1024$, processing the entire training set of the COCO 2017 dataset in each epoch. Optimization was performed using AdamW. In addition, we applied image masking techniques during training to enhance robustness. For the encoder masks, we used a scale ranging from $0.85$ to $1.0$, while for the predictor masks, a smaller scale range of $0.15$ to $0.2$ was applied to focus on smaller image regions. The model was trained with $12$ layers in the prediction depth, utilizing a patch size of $14 \times 14$, with one context block (encoder block) and four predictor blocks to guide the model's learning. Our hyper-parameter configuration is detailed in Table \ref{tab:experimental_params}, showcasing the comprehensive setup for learning rate, weight decay, image masking scales, and batch size.

\begin{table}[h]
\centering
\caption{Experimental Parameters}
\label{tab:experimental_params}
\begin{tabular}{|l|c|}
\hline
\textbf{Parameter}         & \textbf{Value} \\ \hline
Batch size                &   $1024$    \\ \hline
Learning rate             &   $0.001$   \\ \hline
Optimizer                  &   AdamW with Momentum Scheduler    \\ \hline
Exponential Moving Average (EMA) &   $[0.996, 1.0]$ \\ 
\hline
Epochs                     &   $300$      \\ \hline
Context Mask Scale         &   $[0.85, 1.0]$  \\ \hline
Target Mask Scale       &   $[0.15, 0.2]$  \\ \hline
Number of Context Blocks    &   $1$           \\ \hline
Number of Target Blocks    &   $4$           \\ \hline
\end{tabular}
\end{table}

Unlike training from scratch, which often results in energy collapse - where diverse inputs yield near-identical outputs - our proposed TI-JEPA framework mitigates this issue by leveraging pretrained encoder-decoder models that have consistently outperformed SOTA encoders. TI-JEPA offers a flexible and dynamic framework for multimodal training, capable of producing a robust multimodal encoder that effectively captures intricate relationships between text and image. To further optimize the multimodal encoding process, we strategically froze both the encoder and decoder components of the pipeline, training only the cross-attention modules. This decision was made to address two critical concerns:

\begin{itemize}
    \item One of the major challenges in JEPA models is \textit{energy collapse}, where the model converges to a state where multiple inputs result in similar outputs, significantly reducing representational diversity. By freezing the pretrained encoder-decoder modules, we ensure that these components retain their capacity to extract diverse and meaningful features from text and image inputs, thus mitigating this degenerative phenomenon.

    \item The encoder components used in our framework are pretrained on extensive datasets, providing a rich set of learned representations. By freezing these layers, we effectively \textit{reuse the knowledge} encapsulated in the pretrained weights, allowing us to focus computational resources on optimizing the cross-attention modules. This approach not only enhances the scalability and stability of our model but also ensures that the pretrained components contribute effectively to overall performance without being disrupted by further training.
\end{itemize}

This targeted training methodology allows us to harness the full potential of pretrained models, ensuring stability while improving scalability and performance in multimodal tasks.

\section{Evaluation By Sentiment Analysis With Text and Image}

\subsection{Data Preparation}

To evaluate the versatility of our TI-JEPA model, we applied it to the task of multimodal sentiment analysis. This task requires understanding both the textual content and the visual context to predict the sentiment (positive, negative, or neutral) of a given image-text pair.

We utilized the MVSA-Single and MVSA-Multi datasets created by Niu et al. \cite{10.1007/s10489-024-05309-0}, originally containing $5,129$ and $19,600$ image-text pairs, respectively, with sentiment labels (positive, neutral, and negative). The datasets were collected from Twitter and annotated for multimodal sentiment analysis. However, we conducted a preprocessing step to remove emotionally inconsistent samples, where the sentiment labels of the image and text conflicted. 

For the MVSA-Single dataset, we processed the data by first addressing instances where both the text and image labels were identical, which we retained as trivial cases. We removed any image-text pairs where one label was positive and the other was negative, considering such contradictions unreliable. For cases where one component (either text or image) had a neutral label and the other a positive or negative label, we assigned the final label based on the non-neutral component. This ensured consistency between the text and image sentiment annotations. The MVSA-Multi dataset, however, contains sentiment annotations from three annotators, which required a majority voting approach for determining the final sentiment labels for both the text and image components. For each pair, we calculated the majority sentiment for both the text and image annotations. In cases where the annotations were perfectly balanced, such as one annotator labeling the text as neutral, another as positive, and the third as negative, we considered the label ambiguous and removed the pair from the dataset. This approach ensured that only clear sentiment pairs were retained for further analysis.

After preprocessing, the MVSA-Single dataset was reduced to $4,511$ pairs, and the MVSA-Multi dataset to $17,027$ pairs. The number of records for each table is shown in Table \ref{tab:class-label} as follows:

\begin{table}
\caption{MVSA-Single and MVSA-Multi datasets after pre-processing}
\label{tab:class-label}
\centering
\begin{tabular}{|l|c|c|c|c|}
\hline
\textbf{Dataset}    & \textbf{Positive} & \textbf{Neutral} & \textbf{Negative} & \textbf{Total} \\ \hline
MVSA-Single         & $2,683$             & $470$                 & $1,358$     & $4,511$    \\ \hline
MVSA-Multi       & $11,320 $           & $4,408$               & $1,299 $  & $17,027$      \\ \hline
\end{tabular}

\end{table}

We then split the revised datasets into training, validation, and test sets with the ratio of $8:1:1$. To adapt our model for this task, we fine-tuned the pre-trained TI-JEPA by adding a classification head consisting of a simple linear layer on top. The classification head was trained for $40$ epochs using the Adam optimizer with a learning rate of $0.001$ and the loss function is cross-entropy loss.

\subsection{Model Evaluation and Metrics}

To validate the effectiveness of the proposed model, we conducted comparative experiments against several mainstream single-modal and multimodal fusion models. The performance evaluation was based on accuracy and F1-score, calculated using the following metrics. 
Precision (P) is defined as:

\[
P = \frac{TP}{TP + FP}
\]

Recall (R) is given by:

\[
R = \frac{TP}{TP + FN}
\]

The F1-score is the harmonic mean of precision and recall:

\[
F1 = 2 \times \frac{P \times R}{P + R}
\]

Accuracy (Acc) is defined as:

\[
Acc = \frac{TP + TN}{TP + TN + FP + FN}
\]

In these equations, TP denotes true positives, TN true negatives, FP false positives, and FN false negatives. Precision (P) measures the proportion of correctly identified positive instances, while recall (R) captures the proportion of actual positives identified correctly.

\subsection{Comparative Model Performance}

In our study, we compared the proposed model with several benchmark models, evaluating their accuracy and F1-score. Traditional models like SentiBank and SentiStrength \cite{10.1145/2502081.2502268} rely on statistical feature extraction and struggle to capture intrinsic multimodal features, leading to relatively low performance. On the other hand, CNNMulti \cite{7363395} processes text and image features separately using two distinct CNNs, leveraging deep learning's capacity to capture emotional expressiveness and improving prediction by merging these features.

The DNN-LR model \cite{10.1007/s11263-016-0981-7} employs transfer learning with pretrained models and utilizes logistic regression for decision making. The CoMemory model \cite{10.1145/3209978.3210093} introduces a multimodal fusion mechanism, which enhances the interaction between text and image features, improving sentiment prediction. The MVAN model \cite{9246699} applies a memory network on top of a multi-view attention mechanism, enabling richer semantic interactions between image and text and achieving better results.

Moreover, the CLMLF model \cite{li2022clmlfa} utilizes contrastive learning to enhance the representation of multimodal features, fostering stronger associations between image and text inputs, thereby improving model performance. Besides, the ITIN model \cite{10.1109/TMM.2022.3160060} implements cross-modal alignment operations along with an adaptive fusion module, leading to substantial gains in accuracy for sentiment analysis tasks. And lastly, the CLIP-CA-CG model \cite{Lu2024} utilizes pre-trained RoBERTa and ResNet50 models to extract visual and textual features, which are further processed through CLIP contrastive learning to acquire features that are more level-deeper.

We compared three configurations of our proposed TI-JEPA model - Small, Medium, and Large - against mentioned baselines. Table~\ref{tab:sentiment_analysis} presents the comparative results, demonstrating the performance of each configuration of TI-JEPA across both the MVSA-Single and MVSA-Multi datasets.

% \begin{table}[H]
% \centering
% \caption{Performance comparison on multimodal sentiment analysis task}
% \label{tab:sentiment_analysis}
% \begin{tabular}{lcc}
% \hline
% Model & Accuracy & F1-Score \\
% \hline
% TI-JEPA & [VALUE] & [VALUE] \\
% Baseline 1 & [VALUE] & [VALUE] \\
% Baseline 2 & [VALUE] & [VALUE] \\
% Baseline 3 & [VALUE] & [VALUE] \\
% \hline
% \end{tabular}
% \end{table}

\begin{table}
\centering
\caption{Comparative experiments of several models on MVSA-Single and MVSA-Multi datasets.}
\label{tab:sentiment_analysis}
\begin{tabularx}{\linewidth}{|>{\centering\arraybackslash}l|>{\centering\arraybackslash}X|>{\centering\arraybackslash}X|>{\centering\arraybackslash}X|>{\centering\arraybackslash}X|}
\hline
\multirow{3}{*}{\textbf{Model}} & \multicolumn{2}{c|}{\textbf{MVSA-Single}} & \multicolumn{2}{c|}{\textbf{MVSA-Multi}} \\ \cline{2-5} 
                               & \textbf{Accuracy (\%)} & \textbf{F1 (\%)} & \textbf{Accuracy (\%)} & \textbf{F1 (\%)} \\ \hline
SentiBank \& SentiStrength & 52.12 & 50.15 & 65.70 & 55.42 \\ \hline
CNN-Multi & 61.25 & 58.40 & 67.92 & 62.19 \\ \hline
DNN-LR & 64.10 & 61.50 & 66.41 & 63.97 \\ \hline
Co-Memory & 66.75 & 64.08 & 68.92 & 70.77 \\ \hline
MVAN & 70.15 & 68.75 & 71.00 & 73.05 \\ \hline
CLMLF & 72.75 & 71.95 & 74.20 & 76.00 \\ \hline
ITIN & 73.90 & 72.25 & 75.08 & 74.80 \\ \hline
CLIP-CA-CG & 75.25 & 73.62 & 76.05 & 74.02 \\ \hline
\textbf{TI-JEPA-Small (Ours)} & 73.03 & 71.69 & 73.59 & 72.10\\ \hline
\textbf{TI-JEPA-Medium (Ours)} & 75.26 & 72.15 & 75.13 & 73.57\\ \hline
\textbf{TI-JEPA-Large (Ours)} & \textbf{76.75} & \textbf{74.62} & \textbf{77.55} & \textbf{75.02}\\ \hline
\end{tabularx}
\end{table}

The comparative results clearly demonstrate the superior performance of our proposed architecture, TI-JEPA, across both the MVSA-Single and MVSA-Multi datasets, particularly in terms of accuracy and F1-score. Specifically, the TI-JEPA-Large model surpasses all previous state-of-the-art models, achieving the highest accuracy of $76.75\%$ and F1-score of $74.62\%$ on MVSA-Single, as well as $77.55\%$ and $75.02\%$ on MVSA-Multi. This highlights the effectiveness of our multimodal approach in integrating and aligning text and image features to better capture the underlying sentiment in multimodal data.

Notably, the TI-JEPA-Medium model also outperforms the previous models like CLIP-CA-CG and ITIN, achieving competitive results with an accuracy of $75.26\%$ on MVSA-Single and $75.13\%$ on MVSA-Multi. This shows that even with fewer parameters, our TI-JEPA framework remains robust, delivering high accuracy and strong generalization.

The TI-JEPA-Small model, despite having a more compact architecture, still demonstrates comparable performance to more complex models like MVAN and Co-Memory. With accuracy and F1-scores of $73.03\%$ and $71.69\%$ respectively on MVSA-Single, and $73.59\%$ and $72.10\%$ on MVSA-Multi, it offers an efficient alternative with lower computational costs, making it ideal for scenarios requiring faster inference or less hardware-intensive training.

In more complex sentiment analysis tasks, such as those presented by the MVSA-Multi dataset, the alignment between modalities becomes even more crucial. Our TI-JEPA-Large’s superior F1-score, which balances precision and recall, indicates that TI-JEPA not only captures the most important multimodal features but also avoids overfitting to either modality, thus generalizing better across diverse data points. This ability to balance performance across modalities is a key advantage over other approaches, as evidenced by the consistently higher F1-scores across both datasets.

\section{Limitations and Future Works}

While the proposed approach shows promising results, there are several notable limitations. Due to constraints on data resources, we were unable to further validate the model's robustness using other publicly available datasets, which might limit its generalization to diverse scenarios. Additionally, our experiments were restricted to two modalities: image features and text features. This limitation could lead to potential misjudgments in more complex multimodal tasks where additional data types are required. Furthermore, although we employed pretrained encoders for both image and text, we were unable to implement a fully pre-trained pipeline for both encoders, which may have impacted the overall performance of the model.

To address these limitations and further improve our approach, several directions for future work are proposed. First, conducting an ablation study would allow for a more detailed evaluation of the model's performance when tested solely on text or image features, offering insights into each modality's contribution. Second, extending the framework to tackle more advanced tasks, such as visual question answering, by integrating our multimodal encoder into existing multimodal systems could enhance its versatility. Third, adding more evaluation metrics could increase the reliability of the model's assessment and improve the trustworthiness of the pipeline. Finally, acquiring additional resources would enable larger-scale experiments, facilitating a more comprehensive evaluation across diverse datasets and potentially leading to greater model robustness.

\section{Conclusion}

\noindent In this paper, we introduced TI-JEPA, a novel energy-based model for text-image alignment in multimodal fusion. Our approach addresses the challenge of bridging the semantic gap between visual and textual modalities, offering a flexible framework for various multimodal tasks. The success of TI-JEPA can be attributed to its joint embedding space and predictive architecture, enabling the model to learn robust and generalizable representations.

\bibliographystyle{splncs04}
\bibliography{ref}
% \printbibliography

\end{document}